\documentclass[final]{cvpr}

\usepackage{times}
\usepackage{epsfig}
\usepackage{graphicx}
\usepackage{amsmath}
\usepackage{amssymb}

\usepackage{comment}
\usepackage{bbm}
\usepackage{color}
\usepackage{booktabs} 
\usepackage{soul,color}
\usepackage{nicefrac}
\usepackage{subcaption}
\usepackage{multirow}
\usepackage{sidecap}

\usepackage{footnote}
\makesavenoteenv{tabular}
\makesavenoteenv{table}

\usepackage[pagebackref=true,breaklinks=true,colorlinks,bookmarks=false]{hyperref}

\newcommand{\ourmethod}{{STAC}}

\newcommand{\mscoco}{{MS-COCO}}
\newcommand{\pascalvoc}{{PASCAL VOC}}
\newcommand{\voca}{{VOC07}}
\newcommand{\vocb}{{VOC12}}

\def\cvprPaperID{5127} 
\def\confYear{CVPR 2021}

\begin{document}

\title{A Simple Semi-Supervised Learning Framework for Object Detection}

\author{
Kihyuk Sohn\thanks{Equal contribution.} \; Zizhao Zhang\footnotemark[1] \; Chun-Liang Li \; Han Zhang \; Chen-Yu Lee \; Tomas Pfister\\
Google Cloud AI Research, Google Brain\\
{\tt\small\{ksohn,zizhaoz,chunliang,zhanghan,chenyulee,tpfister\}@google.com}
}

\maketitle

\begin{abstract}
Semi-supervised learning (SSL) has a potential to improve the predictive performance of machine learning models using unlabeled data.
Although there has been remarkable recent progress, the scope of demonstration in SSL has mainly been on image classification tasks.
In this paper, we propose {\ourmethod}, a simple yet effective SSL framework for visual object detection along with a data augmentation strategy.
{\ourmethod} deploys highly confident pseudo labels of localized objects from an unlabeled image and updates the model by enforcing consistency via strong augmentations.
We propose experimental protocols to evaluate the performance of semi-supervised object detection using {\mscoco} and show the efficacy of {\ourmethod} on both {\mscoco} and {\voca}. 
On {\voca}, {\ourmethod} improves the AP$^{0.5}$ from $76.30$ to $79.08$; on {\mscoco},
{\ourmethod} demonstrates $2{\times}$ higher data efficiency by achieving 24.38 mAP using only 5\% labeled data than supervised baseline that marks 23.86\% using 10\% labeled data. The code is available at \url{https://github.com/google-research/ssl_detection/}.
\end{abstract}

\vspace{-0.2in}
\section{Introduction}
\label{sec:intro}
\vspace{-0.05in}
Semi-supervised learning (SSL) has received growing attention in recent years as it provides means of using unlabeled data to improve model performance when large-scale annotated data is not available. 
A popular class of SSL methods is based on “Consistency-based Self-Training”~\cite{lee2013pseudo,rasmus2015semi,laine2016temporal,sajjadi2016mutual,tarvainen2017mean,miyato2018virtual,berthelot2019mixmatch,xie2019unsupervised,berthelot2019remixmatch,xie2019self,sohn2020fixmatch}.
The key idea is to first generate the artificial labels for the unlabeled data and train the model to predict these artificial labels when feeding the unlabeled data with semanticity-preserving stochastic augmentations. 
The artificial label can either be a one-hot prediction (hard) or the model's predictive distribution (soft). 
The other pillar for the success of SSL is from advancements in data augmentations. 
Data augmentations improve the robustness of deep neural networks~\cite{simard2003best,krizhevsky2012imagenet} and has been shown to be particularly effective for consistency-based self-training~\cite{xie2019unsupervised,berthelot2019remixmatch,xie2019self,sohn2020fixmatch}. 
The augmentation strategy spans from a manual combination of basic image transformations, such as rotation, translation, flipping, or color jittering, to neural image synthesis~\cite{zhu2017unpaired,hoffman2017cycada,zakharov2019deceptionnet} and policies learned by reinforcement learning~\cite{cubuk2019autoaugment,zoph2019learning}. 
Lately, complex data augmentation strategies, such as RandAugment~\cite{cubuk2019randaugment} or CTAugment~\cite{berthelot2019remixmatch}, have turned out to be powerful for SSL on image classification~\cite{xie2019unsupervised,berthelot2019remixmatch,sohn2020fixmatch,xie2019self}.

\begin{figure}[t]
    \centering
    \includegraphics[width=0.49\textwidth]{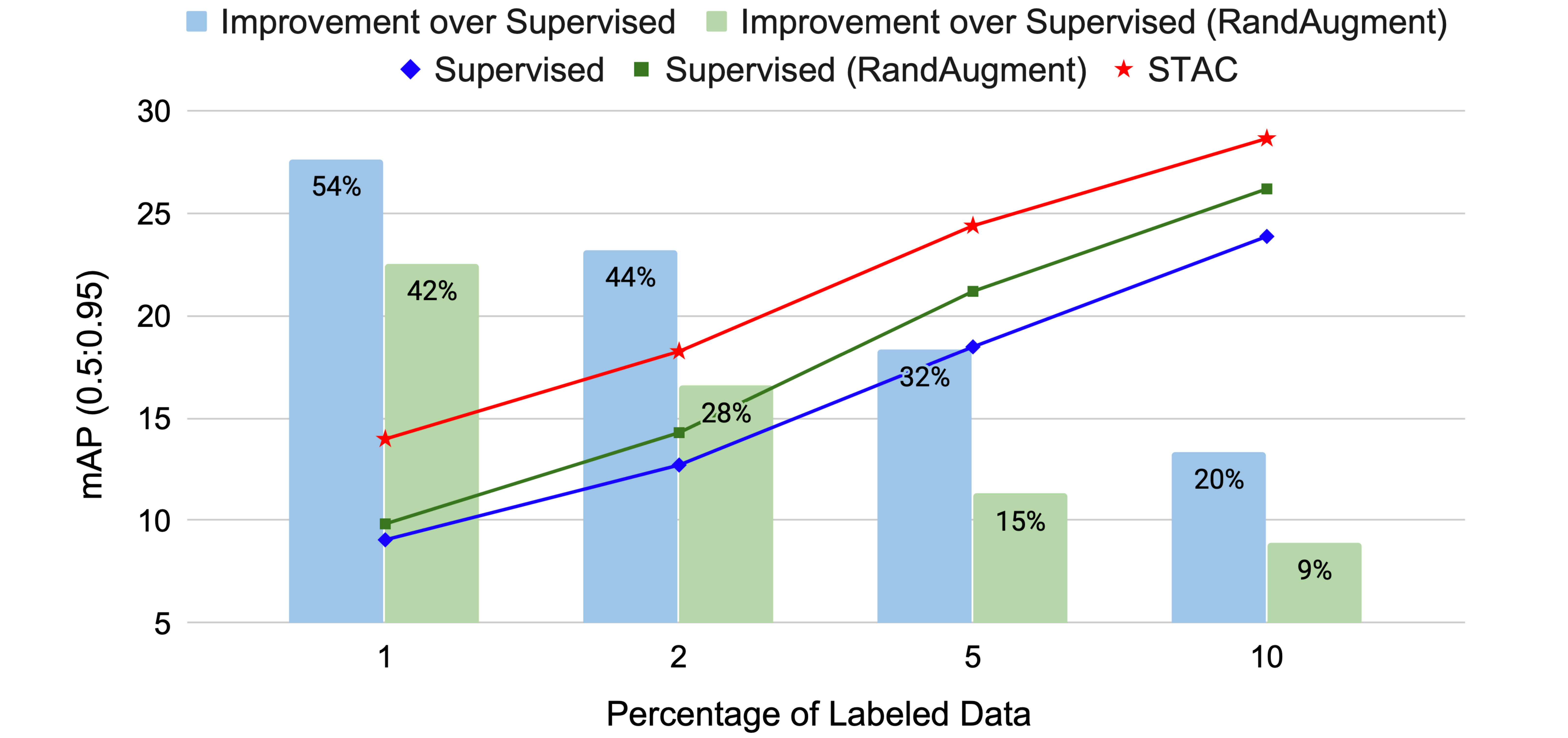}
    \vspace{-0.3in}
    \caption{The proposed semi-supervised learning framework for object detection, {\ourmethod}, consistently improves upon supervised baselines and those with data augmentation using different amount of labeled training data on {\mscoco}~\cite{lin2014microsoft}.}
    \label{fig:teaser}
    \vspace{-0.1in}
\end{figure}

While having made remarkable progress, SSL methods have been mostly applied to image classification, whose labeling cost is relatively cheaper compared to other important problems in computer vision, such as object detection. 
Due to its expensive labeling cost, object detection demands a higher level of label efficiency, necessitating the development of strong SSL methods. 
On the other hand, the majority of existing works on object detection has focused on training a stronger~\cite{shrivastava2016training,lin2017feature,lin2017focal} and faster~\cite{girshick2015fast,ren2015faster,dai2016r} detector given sufficient amount of annotated data. 
Few existing works on SSL for object detection~\cite{tang2016large,misra2015watch,roychowdhury2019automatic} rely on additional context, such as categorical similarities of objects. 

\begin{figure*}[t]
    \centering
    \includegraphics[width=0.7\textwidth]{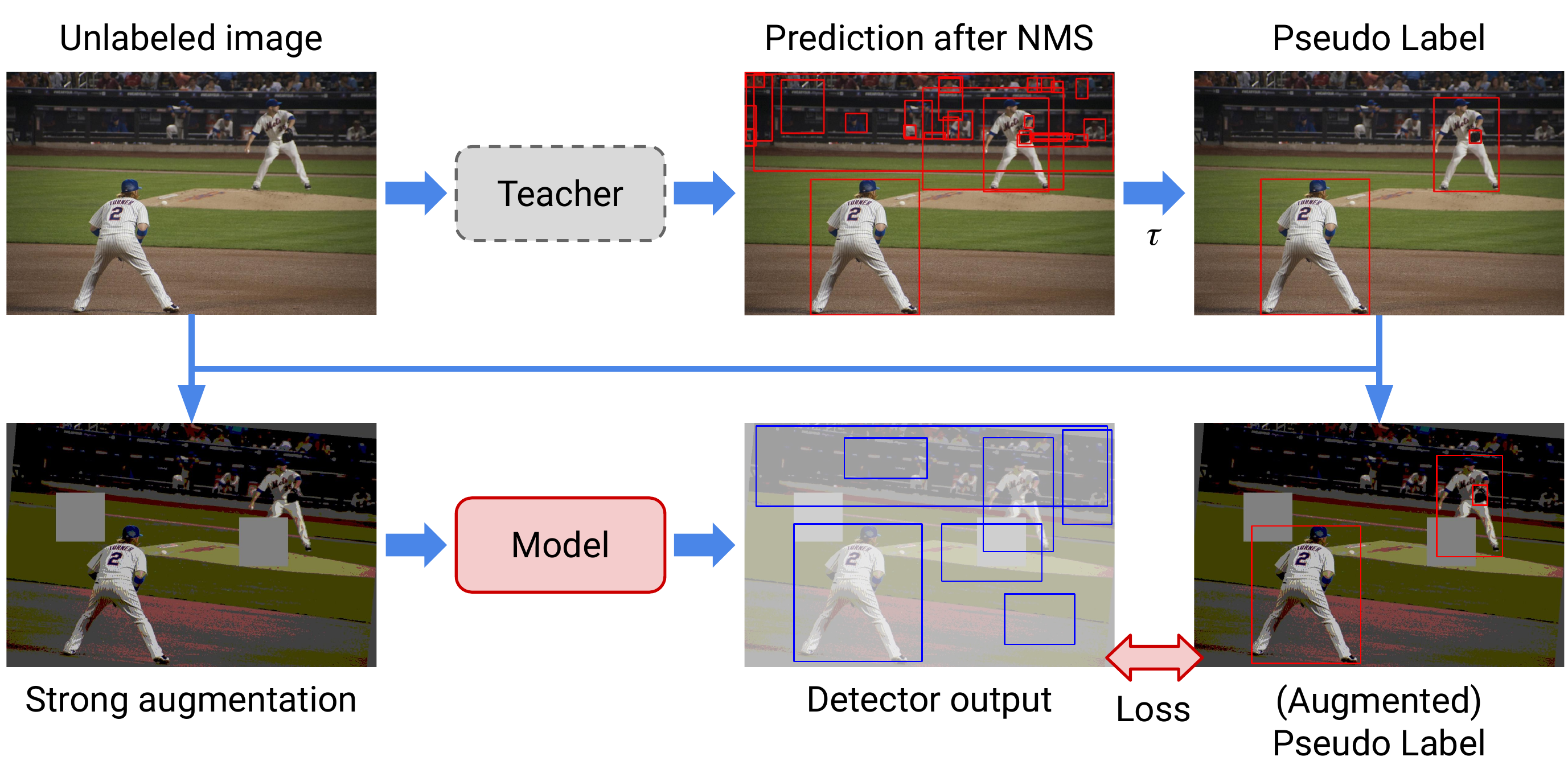}
    \vspace{-0.1in}
    \caption{The proposed SSL framework for object detection. We generate pseudo labels (i.e., bounding boxes and their class labels) for unlabeled data using test-time inference, including NMS \cite{girshick2014rich}, of the teacher model trained with labeled data. We compute unsupervised loss with respect to pseudo labels whose confidence scores are above a threshold $\tau$. The strong augmentations are applied for augmentation consistency during the model training. Target boxes are augmented when global geometric transformations are used.}
    \label{fig:consistency}
    \vspace{-0.1in}
\end{figure*}

In this work, we leverage lessons learned from deep SSL on image classification to tackle SSL for object detection.
To this end, we propose a SSL framework for object detection that combines self-training (via pseudo label)~\cite{scudder1965probability,mclachlan1975iterative} and consistency regularization based on the strong data augmentations~\cite{cubuk2019autoaugment,cubuk2019randaugment,zoph2019learning}.
Inspired by the framework in Noisy-Student~\cite{xie2019self}, our system contains two stages of training. 
In the first stage, we train an object detector (e.g., Faster RCNN~\cite{ren2015faster}) using all labeled data until convergence.
The trained detector is then used to predict bounding boxes and class labels of localized objects for unlabeled images as shown in Figure~\ref{fig:consistency}.
Then, we apply confidence-based filtering to each predicted box (after NMS) with high threshold value to obtain pseudo labels with high precision, inspired by the design of FixMatch~\cite{sohn2020fixmatch}.
In the second stage, the strong data augmentations are applied to each unlabeled image and the model is trained  with labeled data and unlabeled data with its corresponding pseudo labels generated in the first stage.
Encouraged by RandAugment~\cite{cubuk2019randaugment} and its successful adaptation to SSL~\cite{xie2019unsupervised,sohn2020fixmatch} and object detection~\cite{zoph2019learning}, we design our augmentation strategy for object detection, which consists of 
global color transformation, global or box-level~\cite{zoph2019learning} geometric transformations, and Cutout~\cite{devries2017improved}.

We test the efficacy of {\ourmethod} on public datasets: {\mscoco}~\cite{lin2014microsoft} and {\pascalvoc}~\cite{everingham2010pascal}. 
We design new experimental protocols using {\mscoco} dataset to evaluate the semi-supervised object detection performance. 
We use 1, 2, 5 and 10\% of labeled data as labeled sets and the remainder as unlabeled sets to evaluate the effectiveness of SSL methods in the low-label regime. 
In addition, following~\cite{radosavovic2018data,tang2020proposal}, we evaluate using all labeled data as the labeled set and additional unlabeled data provided by {\mscoco} as the unlabeled set. Following~\cite{jeong2019consistency}, we use trainval of {\voca} as the labeled set and that of {\vocb} with or without unlabeled data of {\mscoco} as unlabeled sets. While being simple, {\ourmethod} brings significant gain in mAPs: 18.47 to 24.38 on 5\% protocol, 23.86 to 28.64 on 10\% protocol as in Figure~\ref{fig:teaser}, and 42.60 to 46.01 on {\pascalvoc}.

Overall, the contribution of this paper is as follows:
\begin{enumerate}
    \setlength{\itemsep}{0pt}
    \vspace{-0.1cm}  \item We develop {\ourmethod}, a SSL framework for object detection that seamlessly extends the class of state-of-the-art SSL methods for classification based on self-training and augmentation-driven consistency regularization.
    \vspace{-0.1cm}  \item {\ourmethod} is simple and introduces only two new hyperparameters: the confidence threshold $\tau$ and the unsupervised loss weight $\lambda_u$, which do not require an extensive additional effort for tuning.
    \vspace{-0.1cm}  \item We propose new experimental protocols for SSL object detection using {\mscoco} and demonstrate the efficacy of {\ourmethod} on {\mscoco} and {\pascalvoc} in Faster RCNN framework.
\end{enumerate}

\section{Related Work}
\label{sec:related}
\vspace{-0.05in}

\noindent
\textbf{Object detection} is a fundamental computer vision task and has been extensively studied in the literature~\cite{girshick2014rich,girshick2015fast,ren2015faster,he2017mask,lin2017feature,cai2018cascade,redmon2016you,redmon2017yolo9000,liu2016ssd,lin2017focal}. 
Popular object detection frameworks include Region-based CNN (RCNN) \cite{girshick2014rich,girshick2015fast,ren2015faster,he2017mask,lin2017feature}, YOLO \cite{redmon2016you,redmon2017yolo9000}, SSD \cite{liu2016ssd}, etc \cite{law2018cornernet,tan2019efficientdet,du2019spinenet}. 
The progress made by existing works is mainly on training a stronger or faster object detector given sufficient amount of annotated data.
There is growing interest in improving detectors using unlabeled training data through a semi-supervised object detection framework \cite{tang2016large,misra2015watch,li2020improving}. 
Before deep learning, the idea has been explored by \cite{rosenberg2005semi}. 
Recently, \cite{jeong2019consistency} proposes a consistency-based semi-supervised object detection method, which enforces the consistent prediction of an unlabeled image and its flipped counterpart. 
Their method requires a more sophisticated Jensen-Shannon Divergence for consistency regularization computation.
Similar ideas to consistency regularization have also been studied in the active learning settings for object detection~\cite{wang2018towards}. 
\cite{tang2020proposal} introduces a self-supervised proposal learning module to learn context-aware and noise-robust proposal features from unlabeled data.
\cite{radosavovic2018data} proposes data distillation that generates labels by ensembling predictions of multiple transformations of unlabeled data.
We argue that stronger semi-supervised detectors require further investigation of unsupervised objectives and data augmentations.

\noindent
\textbf{Semi-supervised learning (SSL)} for image classification has been dramatically improved recently. Consistency regularization becomes one of the popular approaches among recent methods \cite{bachman2014learning,sajjadi2016regularization,laine2016temporal,zhai2019s4l,xie2019unsupervised} and inspires \cite{jeong2019consistency} on object detection. The idea is to enforce the model to generate consistent predictions across label-preserving data augmentations. Some exemplars include Mean-Teacher \cite{tarvainen2017mean}, UDA \cite{xie2019unsupervised}, and MixMatch \cite{berthelot2019mixmatch}.  
Another popular class of SSL is pseudo labeling \cite{lee2013pseudo,bachman2014learning}, which can be viewed as a hard version of consistency regularization: the model is performing self-training to generate pseudo labels of unlabeled data and thereby train randomly-augmented unlabeled data to match the respective pseudo labels.
How to use pseudo labels is critical to the success of SSL. 
For instance, Noisy-Student \cite{xie2019self} demonstrates an iterative teacher-student framework that repeats the process of labeling assignments using a teacher model and then training a larger student model.
This method achieves state-of-the-art performance on ImageNet classification by leveraging extra unlabeled images in the wild.
FixMatch~\cite{sohn2020fixmatch} demonstrates a simple algorithm which outperforms previous approaches and establishes state-of-the-art performance, especially on diverse small labeled data regimes.
The key idea behind FixMatch is matching the prediction of the strongly-augmented unlabeled data to the pseudo label of the weakly-augmented counterpart when the model confidence on the weakly-augmented one is high.
In light of the success of these methods, this paper exploits the effective usage of pseudo labeling and pseudo boxes as well as data augmentations to improve object detectors.

\noindent
\textbf{Data augmentations} are critical to improve model generalization and robustness~\cite{cubuk2019autoaugment,cubuk2019randaugment,hendrycks2019augmix,zoph2019learning,zhang2017mixup,zhong2017random,devries2017improved,dwibedi2017cut,ho2019population}, especially gradually become a major impetus on semi-supervised learning ~\cite{berthelot2019mixmatch,berthelot2019remixmatch,xie2019unsupervised,sohn2020fixmatch}.
Finding appropriate color transformations and geometric transformations of input spaces has been shown to be critical to improve  generalization \cite{cubuk2019autoaugment,hendrycks2019augmix}. 
However, most augmentations are mainly studied in image classification. The complexity of data augmentations for object detection is much higher than image classification \cite{zoph2019learning}, since global geometric transformations of data affect bounding box annotations. 
Some works have presented augmentation techniques for supervised object detection, such as MixUp~\cite{zhang2017mixup,zhang2019bag}, CutMix~\cite{yun2019cutmix}, or augmentation strategy learning \cite{zoph2019learning}.
The recent consistency-based SSL object detection method~\cite{jeong2019consistency} utilizes global horizontal flipping (weak augmentation) to construct the consistency loss. To the best of our knowledge, the impact of intensive data augmentations on semi-supervised object detection has not been thoroughly studied.

\section{Methodology}
\label{sec:method}
\vspace{-0.05in}

\subsection{Background: Unsupervised Loss in SSL}
\label{sec:prelim_fixmatch}
\vspace{-0.03in}
Formulating an unsupervised loss that leverages unlabeled data is the key in SSL. Many advancements in SSL for classification rely on some forms of \textit{consistency regularization}~\cite{lee2013pseudo,rasmus2015semi,laine2016temporal,sajjadi2016mutual,tarvainen2017mean,miyato2018virtual,berthelot2019mixmatch,xie2019unsupervised,berthelot2019remixmatch,xie2019self}. Inspired by a comparison in~\cite{sohn2020fixmatch}, we provide a unified view of consistency regularization for image classification.
For $K$-way classification, the consistency regularization is written as follows:
\begin{equation}
    \ell_{u} = \sum_{x\in\mathcal{X}} w(x)\, \ell\left(q(x), p(x;\theta)\right), \label{eq:unsup_loss_general}
\end{equation}
where $x\,{\in}\,\mathcal{X}$ is an image, $p,q\,{:}\,\mathcal{X}\,{\rightarrow}\,[0,1]^{K}$ map  $x$ into a $(K{-}1)$-simplex, and $w\,{:}\,\mathcal{X}\,{\rightarrow}\,\{0,1\}$ maps $x$ into a binary value. $\ell(\cdot,\cdot)$ measures a distance between two vectors. Typical choices include $L_2$ distance and cross entropy. Here, $p$ represents the prediction of the model parameterized by $\theta$, $q$ is the prediction target, and $w$ is the weight that determines the contribution of $x$ to the loss. As an example, pseudo labeling~\cite{lee2013pseudo} has the following configurations:
\begin{align}
    q(x)\,&{=}\,\mbox{\textsc{one\_hot}}(\arg\max\left(p(x;\theta\right))) \nonumber \\
    w(x)\,&{=}\,\mathbbm{1} \, \mbox{ if }\max(p(x;\theta))\geq\tau
    \label{eq:pseudo_label}
\end{align}
We refer to Appendix for configurations of SSL methods.
State-of-the-art methods, such as Unsupervised Data Augmentation~\cite{xie2019unsupervised} and FixMatch~\cite{sohn2020fixmatch}, apply strong data augmentation $\mathcal{A}$, such as RandAugment~\cite{cubuk2019randaugment} or CTAugment~\cite{berthelot2019remixmatch}, to the model prediction $p(\mathcal{A}(x);\theta)$ for improved robustness. Noisy-Student~\cite{xie2019self} applies diverse forms of stochastic noise to the model prediction, including input augmentations via RandAugment, and network augmentations via dropout~\cite{srivastava2014dropout} and stochastic depth~\cite{huang2016deep}. While sharing similarities on the model prediction, they differ in $q$ that generates the prediction target as detailed in Appendix. Different from Equation~\eqref{eq:pseudo_label} and many aforementioned algorithms,
Noisy-Student employs a ``teacher'' network other than $p(\cdot,\theta)$ to generates pseudo labels $q(x)$. Note that the fixed teacher network allows offline pseudo label generation and this provides scalability to large unlabeled data and flexibility on the choice of architecture or optimization.

\subsection{{\ourmethod}: SSL for Object Detection}
\label{sec:detmatch_algo}
\vspace{-0.03in}
We propose a simple SSL framework for object detection, called \textbf{{\ourmethod}}, based on the \textbf{S}elf-\textbf{T}raining (via pseudo label) and the \textbf{A}ugmentation driven \textbf{C}onsistency regularization. First, we adopt a stage-wise training of Noisy-Student~\cite{xie2019self} for its scalability and flexibility. This involves at least two stages of training, where in the first stage, we train a teacher model using all available labeled data, and in the second stage, we train {\ourmethod} using both labeled and unlabeled data. Second, we use a high threshold value for the confidence-based thresholding inspired by FixMatch~\cite{sohn2020fixmatch} to control the quality of pseudo labels comprised of bounding boxes and their class labels in object detection. The steps for training {\ourmethod} are summarized as follows:
\begin{enumerate}
    \setlength{\itemsep}{0pt}
    \vspace{-0.1cm}  \item \textbf{Train a teacher model} on available labeled images.
    \vspace{-0.1cm}  \item \textbf{Generate pseudo labels} of unlabeled images (i.e., bounding boxes and their class labels) using the trained teacher model.
    \vspace{-0.1cm}  \item \textbf{Apply strong data augmentations} to unlabeled images, and augment pseudo labels (i.e. bounding boxes) correspondingly when global geometric transformations are applied.
    \vspace{-0.1cm}  \item \textbf{Compute unsupervised loss} and supervised loss to train a detector.
\end{enumerate}
\textbf{Training a Teacher Model.}
We develop our formulation based on the Faster RCNN~\cite{ren2015faster} as it has been one of the most representative detection framework. 
Faster RCNN has a classifier (CLS) and a region proposal network (RPN) heads on top of the shared backbone network. Each head has two modules, namely region classifiers (e.g., a ($K{+}1$)-way classifier for the CLS head or a binary classifier for the RPN head) and bounding box regressors (REG). 
We present the supervised and unsupervised losses of the Faster RCNN for the RPN head for simplicity.
The supervised loss is written as follows:
%
\begin{align}
    & \ell_s(x,\mathbf{p}^{*},\mathbf{t}^{*}) =  \sum_{i}\ell_{s}\big(x,p_{i}^{*},t_i^{*}\big) \nonumber\\
    & = \sum_{i}\Big[\frac{1}{N_{\text{cls}}}\mathcal{L}_{\text{cls}}\left(p_{i}, p_{i}^{*}\right) + \frac{\lambda}{N_{\text{reg}}}\mathcal{L}_{\text{reg}}\left(t_{i}, t_{i}^{*}\right)\Big]
    \label{eq:sup_loss}
\end{align}
where $i$ is an index of an anchor in mini-batch. $p_i$ is the predictive probability of an anchor being positive, $t_i$ is the 4-dimensional coordinates of an anchor. $p_{i}^{*}$ is the binary label of an anchor with respect to ground-truth boxes,  $t_{i}^{*}$ is the ground-truth box coordinates of the box $i$ for all $p_i^* = 1$. 

\noindent
\textbf{Generating Pseudo Labels.}
We perform a test-time inference of the object detector from the teacher model to generate pseudo labels. That being said, the pseudo label generation involves not only the forward pass of the backbone, RPN and CLS networks, but also the post-processing such as non-maximum suppression (NMS). This is different from conventional approaches for classification where the confidence score is computed from the raw predictive probability. We use the score of each returned bounding box after NMS, which aggregates the prediction probabilities of anchor boxes. Using box predictions after NMS has an advantage over using raw predictions (before NMS) since it removes repetitive detection. However, this does not filter out boxes at wrong locations as visualized in Figure~\ref{fig:consistency} and Figure~\ref{fig:exp_ablation_threshold_0_0}. We apply confidence-based thresholding~\cite{xie2019self,sohn2020fixmatch} to further reduce potentially wrong pseudo boxes.

\noindent
\textbf{Unsupervised Loss.}
Given an unlabeled image $x$, a set of predicted bounding boxes and their region proposal confidence scores, we determine $q_{i}^{*}$, a binary label of an anchor $i$ with respect to pseudo boxes, for all anchors. 
Note that \emph{the simple threshold mechanism $w$ in Equation \eqref{eq:pseudo_label} is applied on $q_{i}^{*}$ using the CLS head, so it is 1 if anchor is associated with any pseudo boxes whose CLS prediction confidence scores of teacher model are higher than the threshold $\tau$ and $0$ otherwise (i.e. treated as background)}.
Let $\mathbf{s}^{*}$ be box coordinates of pseudo boxes. 
Then, the unsupervised RPN loss of {\ourmethod} is written as $\ell_u(\mathcal{A}(x_u,\mathbf{s}^{*}), \mathbf{q}^{*}) \,{=}\, \ell_s(x_{\mathcal{A}}, \mathbf{q}^{*}, \mathbf{s}^{*}_\mathcal{A})$, where $\mathcal{A}$ is a strong data augmentation applied to an unlabeled image $x$, yielding $x_{\mathcal{A}}$. Since some transformation operations are not invariant to the box coordinates (e.g., global geometric transformation~\cite{zoph2019learning}), the operations $\mathcal{A}$ is applied on the pseudo box coordinates as well, yielding $\mathbf{s}^{*}_{\mathcal{A}}$.

Finally, the RPN is trained by jointly minimizing two losses as follows:
\begin{equation}
    \ell = \ell_{s}(x_s,\mathbf{p}^{*}, \mathbf{t}^{*}) + \lambda_u \ell_{u}(\mathcal{A}(x_u,\mathbf{s}^{*}),\mathbf{q}^{*}).
\end{equation}
{\ourmethod} introduces two hyperparameters $\tau$ and $\lambda_u$. In experiments, we find $\tau\,{=}\,0.9$ and $\lambda_u\,{\in}\,[1,2]$ work well.
Note that the consistency-based SSL object detection method in~\cite{jeong2019consistency} requires sophisticated weighting schedule for $\lambda_u$ including temporal ramp-up and ramp-down. Instead, our framework demonstrates effectiveness with a simple constant schedule thanks to the consistency regularization using a strong data augmentation and confidence-based thresholding.

\begin{figure}[t]
    \centering
    \includegraphics[width=0.495\textwidth]{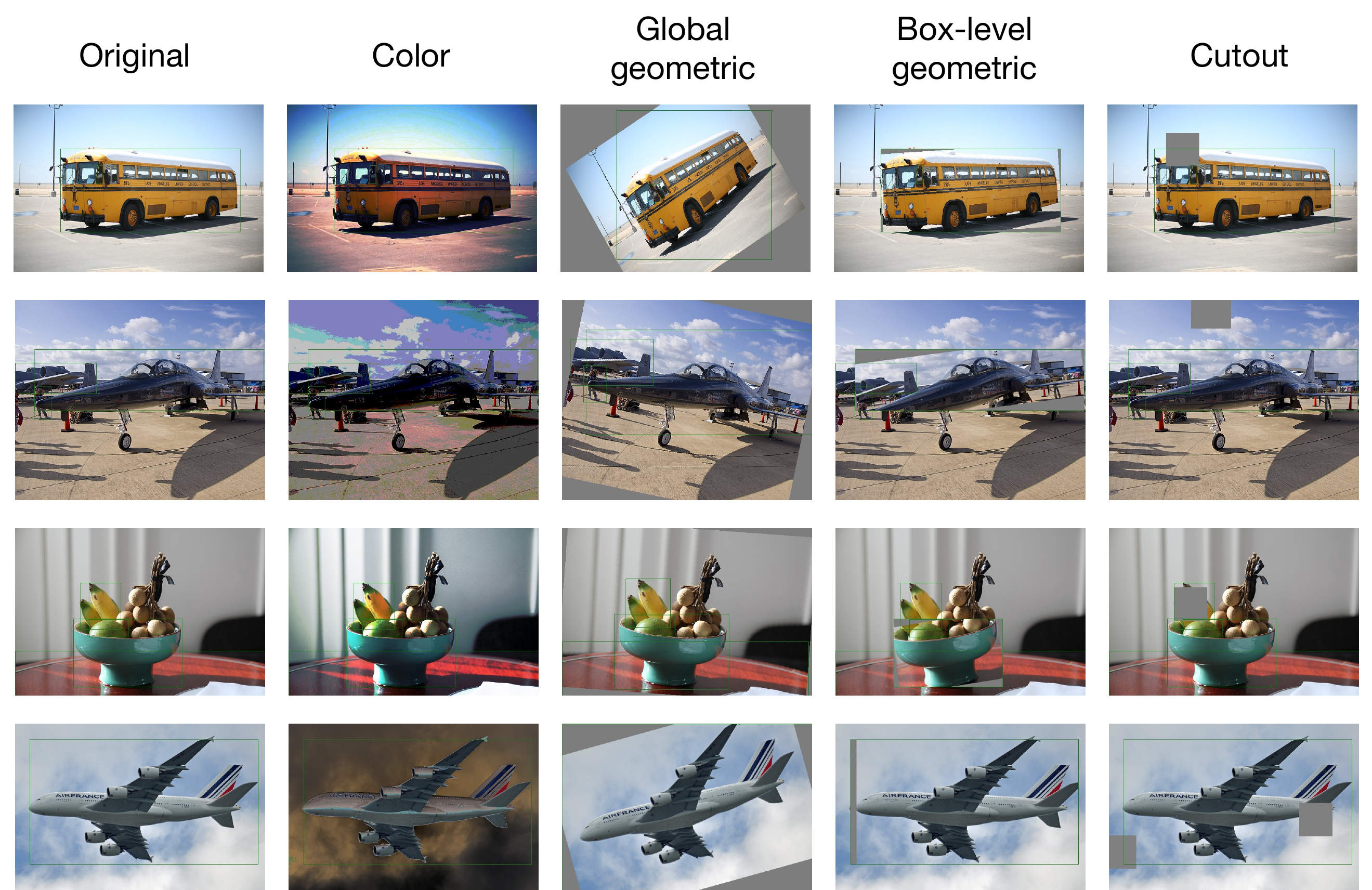}
    \vspace{-0.25in}
    \caption{Visualization of different types of augmentation strategies. From left to right: original image, color transformation, global geometric transformation, box-level geometric transformation, box-level geometric transformation, and Cutout. }
    \label{fig:visualize_augmentation}
\end{figure}

\begin{table*}[t]
    \centering
    \begin{tabular}{l||c|c|c|c|c}
        \toprule
        Methods & { }1\% COCO{ } & { }2\% COCO{ } & { }5\% COCO{ } & { }10\% COCO{ } & { }100\% COCO{ } \\
        \midrule
        {}Supervised{ } & 9.05$\pm$0.16 & 12.70$\pm$0.15 & 18.47$\pm$0.22 & 23.86$\pm$0.81 & 37.63  \\
        {}Supervised$^{\dagger}${ } & 9.83$\pm$0.23 & 14.28$\pm$0.22 & 21.18$\pm$0.20 & 26.18$\pm$0.12 & \textbf{39.48} 	 \\
        {}{\ourmethod}{ } & \textbf{13.97}$\pm$0.35 & \textbf{18.25}$\pm$0.25 & \textbf{24.38}$\pm$0.12 & \textbf{28.64}$\pm$0.21 & 39.21  \\
        \bottomrule
    \end{tabular}
    \vspace{-0.1in}
    \caption{Comparison in mAPs for different methods on {\mscoco}. We report the mean and standard deviation over 5 data folds for 1, 2, 5 and 10\% protocols. ``Supervised'' refers to models trained on labeled data only, which then are used to provide pseudo labels for {\ourmethod}. We train {\ourmethod} with the \textsf{C+\{B,G\}+Cutout} augmentation for unlabeled data. Models with $^{\dagger}$ are trained with the same augmentation strategy, but only with labeled data. See Section~\ref{sec:exp_main} for more details.}
    \label{tab:main_result}
    \vspace{-0.1in}
\end{table*}

\noindent
\textbf{Data Augmentation Strategy.} 
The key factor for the success of consistency-based SSL methods, such as UDA~\cite{xie2019unsupervised} and FixMatch~\cite{sohn2020fixmatch}, is a strong data augmentation. While the augmentation strategy for supervised and semi-supervised image classification has been extensively studied~\cite{cubuk2019autoaugment,cubuk2019randaugment,berthelot2019remixmatch,xie2019unsupervised,sohn2020fixmatch}, not much effort has been made yet for object detection. We extend the RandAugment for object detection used in~\cite{cubuk2019autoaugment} using the augmentation search space recently proposed by~\cite{zoph2019learning} (e.g., box-level transformation) along with the Cutout~\cite{devries2017improved}.
We explore different variants of transformation operations and determinate a group of effective combinations. Each operation has a magnitude that decides the augmentation degree of strength.\footnote{The range of degrees is empirically chosen without tuning.}
\begin{enumerate}
    \setlength{\itemsep}{0pt}
    \vspace{-0.1cm} \item {Global color transformation (\textsf{C})}: Color transformation operations in~\cite{cubuk2019randaugment} and the suggested ranges of magnitude for each op are used. 
    \vspace{-0.1cm}  \item {Global geometric transformation (\textsf{G})}: Geometric transformation operations in~\cite{cubuk2019randaugment}, namely, \textit{x-y} translation, rotation, and \textit{x-y} shear, are used.\footnote{The translation range in percentage is [$-10\%$, $10\%$] of image widths or heights. The rotation and shear ranges are [$-30\%$, $30\%$] in degrees.}
    \vspace{-0.1cm} \item {Box-level transformation~\cite{zoph2019learning} (\textsf{B})}: Three transformation operations from global geometric transformations are used, but with smaller magnitude ranges.\footnote{The translation range in percentage is [$-5\%$, $5\%$] of image widths or heights. The rotation and shear range is [$-10\%$, $10\%$] in degree.}
\end{enumerate}
For each image, we apply transformation operations in sequence as follows. First, we apply one of the operations sampled from \textsf{C}. Second, we apply one of the operations sampled from either \textsf{G} or \textsf{B}. Finally, we apply Cutout at multiple random locations\footnote{The number of Cutout regions is sampled from [1, 5], and the region size is sampled from [0\%, 20\%] of the short edge of the applied image.} of a whole image to prevent a trivial solution when applied exclusively inside the bounding box. We visualize transformed images with aforementioned augmentation strategies in Figure~\ref{fig:visualize_augmentation}.

\begin{table}[t]
    \centering
    \begin{tabular}{l||c|c}
        \toprule
        Methods & { } mAP &  { } AP$^{0.5}$\\
        \midrule
        {}Supervised{ } & 42.60 & 76.30 \\
        {}Supervised$^{\dagger}${ } & 43.40 & 78.21 \\
        {}{\ourmethod} (+\vocb){ } &  44.64 & 77.45 \\
        {}{\ourmethod} (+\vocb{ }\& COCO){ } &  \textbf{46.01} & \textbf{79.08} \\
        \midrule
        \cite{jeong2019consistency} (+\vocb{ }\& COCO) { } & - & 75.1\footnote{We note that the number from \cite{jeong2019consistency} using ResNet101 with R-FCN while all the results from our implementation use ResNet50 with FPN.} \\ 
        \bottomrule
    \end{tabular}
    \vspace{-0.05in}
    \caption{Comparison in mAPs for different methods on {\voca}. We report both mAPs at IoU=$0.5{:}0.95$, a standard metric for {\mscoco}, as well as at IoU=$0.5$ (AP$^{0.5}$), since AP$^{0.5}$ is a saturated metric as pointed out by~\cite{cai2018cascade}. For {\ourmethod}, we follow~\cite{jeong2019consistency} to have different level of unlabeled sources, including {\vocb} and the subset of {\mscoco} data with the same classes as {\pascalvoc}.}
    \label{tab:voc_result}
    \vspace{-0.2in}
\end{table}

\begin{table*}[t]
    \centering
    \begin{tabular}{l|c|c|c|c}
        \toprule
        Augmentation & -- & \textsf{C} & { }\textsf{C+\{G,B\}}{ } & { }\textsf{C+\{G,B\}+Cutout}{} \\
        \midrule
        5\% {\mscoco} (\textsf{quick}) & { }18.67{ } & { }20.13{ } & { }20.78{ } & { }\textbf{21.16}{ } \\
        10\% {\mscoco} (\textsf{quick}) & { }24.05{ } & { }25.26{ } & { }25.92{ } & { }\textbf{26.34}{ } \\
        10\% {\mscoco} (\textsf{standard}) & { }19.74{ } & { }21.40{ } & { }24.24{ } & { }\textbf{24.65}{ } \\
        100\% {\mscoco} (\textsf{standard}) & { }\textbf{37.42}{ } & { }37.22{ } & { }36.39{ } & { }36.12{ } \\
        100\% {\mscoco} (\textsf{standard}, $2{\times}$){ } & { }37.88{ } & { }\textbf{38.91}{ } & { }38.73{ } & { }38.57{ } \\
        100\% {\mscoco} (\textsf{standard}, $3{\times}$){ } & { }37.63{ } & { }39.33{ } & { }\textbf{39.75}{ } & { }39.48{ } \\
        \bottomrule
    \end{tabular}
    \vspace{-0.1in}
    \caption{mAPs of supervised models trained with different augmentation and learning schedules. We test on a single fold of 5\% and 10\% protocols. See Section~\ref{sec:exp_ablation_augmentation} for more details. Bold text indicates the best number in each row.}
    \label{tab:ablation_augmentation}
    \vspace{-0.1in}
\end{table*}

\section{Experiments}
\label{sec:exp}
\vspace{-0.05in}
We test the efficacy of our proposed method on {\mscoco}~\cite{lin2014microsoft}, which is one of the most popular public benchmarks for object detection. 
{\mscoco} contains more than 118k labeled images and 850k labeled object instances from 80 object categories for training. In addition, there are 123k unlabeled images that can be used for semi-supervised learning. 
We experiment two SSL settings. First, we randomly sample 1, 2, 5 and 10\% of labeled training data as a labeled set and use the rest of labeled training data as an unlabeled set. For these experiments, we create 5 data folds. 1\% protocol contains approximately 1.2k labeled images randomly selected from the labeled set of {\mscoco}. 2\% protocol contains additional $\sim$1.2k images and 5, 10\% protocol datasets are constructed in a similar way.
Second, following~\cite{tang2020proposal}, we use an entire labeled training data as a labeled set and additional unlabeled data as an unlabeled set. Note that the first protocol tests the efficacy of {\ourmethod} when only few labeled examples are available, while the second protocol evaluates the potential to improve the state-of-the-art object detector with unlabeled data in addition to already a large-scale labeled data.
We report the mAP over 80 classes. 

We also test on {\pascalvoc}~\cite{everingham2010pascal} following~\cite{jeong2019consistency}. The trainval set of {\voca}, containing 5,011 images from 20 object categories, is used as a labeled training data, and 11,540 images from the trainval set of {\vocb} are used for an unlabeled training data. The detection performance is evaluated on the test set of {\voca} and mAP at IoU of $0.5$ (AP$^{0.5}$) is reported in addition to the {\mscoco} metric. 

\subsection{Implementation Details}
\label{sec:exp_impl_detail}
\vspace{-0.03in}
Our implementation is based on the Faster RCNN and FPN library of Tensorpack~\cite{wu2016tensorpack}. 
We use ResNet-50~\cite{he2016deep} backbone for our object detector models. Unless otherwise stated, the network weights are initialized by the ImageNet-pretrained model at all stages of training.

Since the training of the object detector is quite involved, we stay with the default learning settings for all our experiments other than the learning schedule.
Most of our experiments are conducted using the \textsf{quick} learning schedule\footnote{Section~\ref{sec:exp_ablation_augmentation} defines different learning schedules.} with an exception for 100\% {\mscoco} protocol.\footnote{\scriptsize\url{https://github.com/tensorpack/tensorpack/tree/master/examples/FasterRCNN\#results}} 
We find that the model's performance is benefited significantly by longer training when more labeled training data and more complex data augmentation strategies are used. {\ourmethod} introduces two new hyperparameters $\tau$ for the confidence threshold and $\lambda_u$ for the unsupervised loss. We use $\tau\,{=}\,0.9$ and $\lambda_u\,{=}\,2$ for all experiments except for the 100\% protocol of {\mscoco}, were we lower threshold $\tau\,{=}\,0.5$ to increase the recall of pseudo labels. We refer readers to Appendix for complete learning settings.

\begin{table*}[t]
    \centering
    \begin{tabular}{l||c|c||c|c|c|c|c}
        \toprule
        Unlab. Size{ } & Sup. & Sup.$^{\dagger}$ & 1${\times}$ & 2${\times}$ & 4${\times}$ & 8${\times}$ & { }Full \\
        \midrule
        5\% {\mscoco}{ } & { }18.67{ } & { }21.16{ } & { }19.81{ } & { }20.79{ } & { }22.09{ } & { }23.14{ } & { }24.49{ } \\
        10\% {\mscoco}{ } & { }24.05{ } & { }26.34{ } & { }25.38{ } & { }26.52{ } & { }27.33{ } & { }27.95{ } & { }29.00{ } \\
        \bottomrule
    \end{tabular}
    \vspace{-0.1in}
    \caption{mAPs of {\ourmethod} trained with varying amount of unlabeled data. $[n]{\times}$ refers that the amount of unlabeled data is $[n]$ times larger than that of labeled data. We test on a single fold of 5\% and 10\% protocols.}
    \label{tab:ablation_unlab}
    \vspace{-0.1in}
\end{table*}

\subsection{Results}
\label{sec:exp_main}
\vspace{-0.03in}
Since deep semi-supervised learning of visual object detectors has not been widely studied yet, we mainly compare {\ourmethod} with the supervised models (i.e., models trained with labeled data only) for various experimental protocols using different data augmentation strategies. Table~\ref{tab:main_result} summarizes the results. For 1, 2, 5 and 10\% protocols, we train models with a \textsf{quick} learning schedule and report mAPs averaged over 5 data folds and their standard deviation. For 100\% protocol, we employ \textsf{standard} with $3{\times}$ longer learning schedule and report a single mAP value for each model.

Firstly, we confirm the findings of~\cite{cubuk2019randaugment} with varying amount of labeled training data that the RandAugment improves the supervised learning performance of a detector by a significant margin, 2.71 mAP at 5\% protocol, 2.32 mAP at 10\% protocol, and 1.85 mAP for 100\% protocol, upon the supervised baselines with default data augmentation of resizing and horizontal flipping.

{\ourmethod} further improves the performance upon stronger supervised models. We find it to be particularly effective for protocols with small labeled training data, showing 5.91 mAP improvement at 5\% protocol and 4.78 mAP at 10\% protocol. Interestingly, {\ourmethod} is proven to be at least 2${\times}$ more data efficient than the baseline models for both 5\% (24.36 for {\ourmethod} v.s. 23.86 for supervised model with 10\% labeled training data) and 10\% protocols (28.56 for {\ourmethod} v.s. 28.63 for the supervised model with 20\% labeled training data). For the 100\% protocol, {\ourmethod} achieves 39.21 mAP. This improves upon the baseline (37.63 mAP), but falls short of the supervised model with a strong data augmentation (39.48 mAP). We hypothesize that the pseudo label training benefits by a larger amount of unlabeled data relative to the size of labeled data and study its effectiveness with respect to the scale of unlabeled data in Section~\ref{sec:exp_ablation}.

We have a similar finding for experiments on {\pascalvoc}.
In Table~\ref{tab:voc_result}, the mAP of the supervised models increases from 42.6 to 43.4, and AP$^{0.5}$ increases from 76.30 to 78.21.  A large-scale unlabeled data from {\vocb} and {\mscoco} further improves the performance, achieving 46.01 mAP and 79.08 AP$^{0.5}$.

\begin{figure}[t!]
    \centering
    \includegraphics[width=0.5\textwidth]{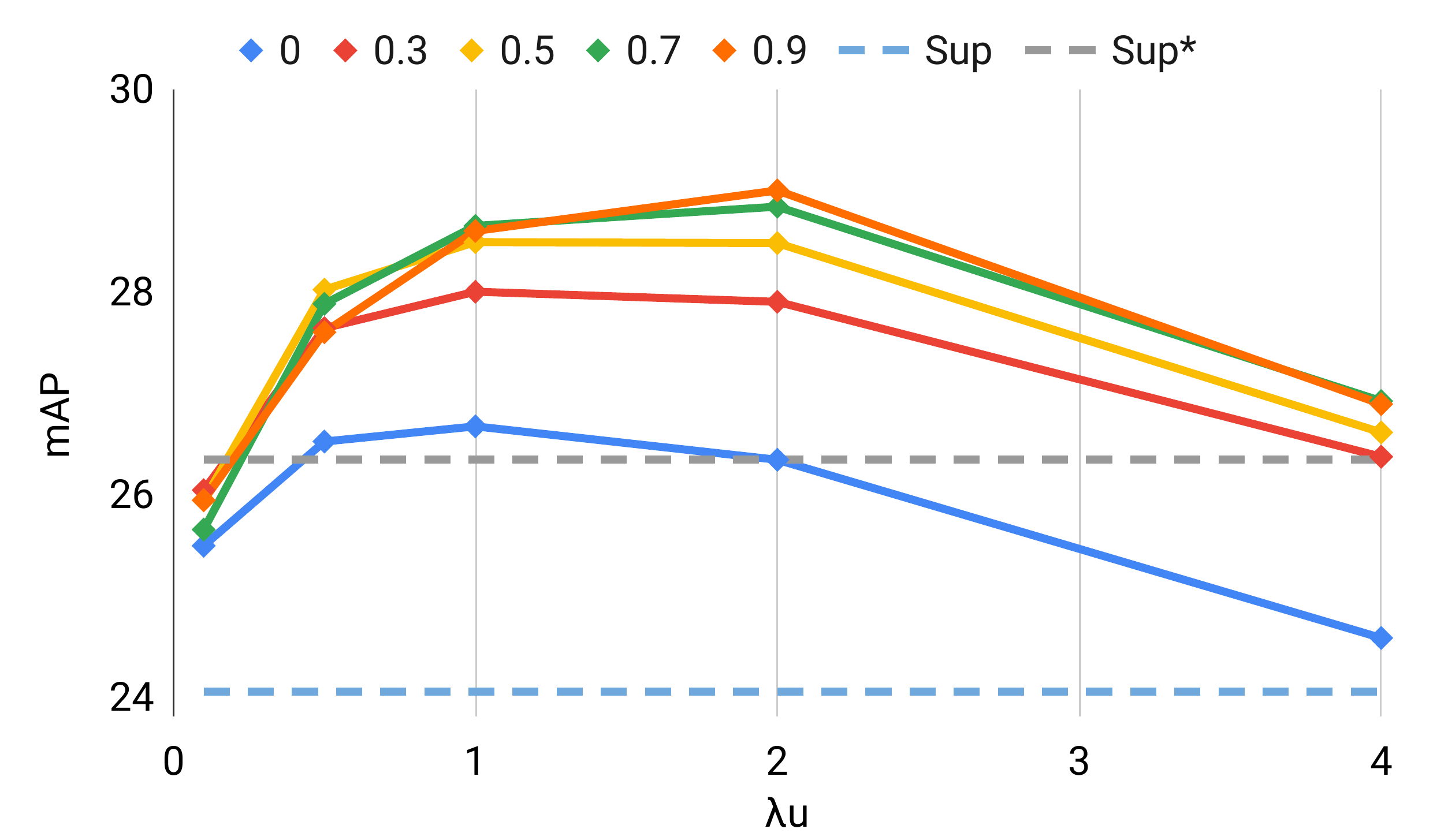} \vspace{-.6cm}
    \caption{mAPs of {\ourmethod} with different values of $\lambda_u\,{\in}\,\{0.1,0.5,1,2,4\}$ and $\tau\,{\in}\,\{0,0.3,0.5,0.7,0.9\}$. We test on a single fold of 10\% protocol. Different colors represent mAPs of models with different $\tau$ values. ``Sup'' represents the mAP of supervised model with default augmentations and ``Sup*'' represents that with \textsf{C+\{G,B\}+Cutout}.}
    \label{fig:exp_ablation_hparams}
    \vspace{-.5cm}
\end{figure}

\section{Ablation Study}
\label{sec:exp_ablation}
\vspace{-0.05in}
We perform ablation study on the key components of {\ourmethod}. The study analyzes the impact on the detector performance of 1) different data augmentations and learning schedule strategies, 2) different sizes of unlabeled sets, 3) the hyperparameters $\lambda_u$, coefficient for unsupervised loss, and $\tau$, confidence threshold, and 4) quality of pseudo labels and their impact on the proposed {\ourmethod}.

\subsection{Data Augmentation and Learning Schedule}
\label{sec:exp_ablation_augmentation}
\vspace{-0.03in}
In this section, we evaluate the performance of supervised detector models with different data augmentation strategies and learning rate schedules while varying the amount of training data. 
We consider different combinations of augmentation modules, including the default augmentations of horizontal image flip, color only (\textsf{C}), color followed by geometric or box-level transforms (\textsf{C+\{G,B\}}), and the one followed by Cutout (\textsf{C+\{G,B\}+Cutout}). For \textsf{\{G,B\}}, we sample randomly and uniformly between geometric and box-level transform modules for each image.
We consider different learning schedules, including \textsf{quick}, \textsf{standard}, and \textsf{standard $[n]{\times}$} (standard setting with $[n]$ times longer training). While the number of weight updates are the same, the \textsf{quick} schedule uses lower resolution image as an input and smaller batch size for training.

The summary results are provided in Table~\ref{tab:ablation_augmentation}.
With small amount of labeled training data, we observe an increasing positive impact on detector performance with more complex (thus stronger) augmentation strategies. The trend holds true with the \textsf{standard} schedule, but we find that the \textsf{quick} schedule is beneficial on the low-labeled data regime due to its fast training and less issue of overfitting.
On the other hand, we observe that the network significantly underfits with our augmentation strategies when all labeled data is used for training. For example, with 100\% labeled data, we achieve even lower mAP of $36.12$ with \textsf{C+\{G,B\}+Cutout} strategy than that of $37.42$ with default augmentations. We find that the issue can be alleviated by longer training. Moreover, while the performance with default augmentations saturates and starts to decrease as it is trained longer, the models with strong data augmentation start to outperform, demonstrating their effectiveness on training with large-scale labeled data.

{\ourmethod} contains two key components: self-training and strong data augmentation.
We also verify the importance of data augmentation in Appendix, which is in line with recent findings in SSL for image classification~\cite{sohn2020fixmatch}. We evaluate the performance of {\ourmethod} with the default augmentations (horizontal flip). On a single fold of 10\% protocol, we observe a good improvement in mAP upon baseline model (from 24.05 to 26.27), but the gain is not as significant as {\ourmethod} (29.00). On 100\% protocol, we observe slight decrease in performance when trained with self-training only (from 37.63 to 37.57), while {\ourmethod} achieves 39.21 in mAP.

\begin{figure}[t]
    \centering
    \begin{subfigure}{.08\textwidth}
        \includegraphics[width=\textwidth]{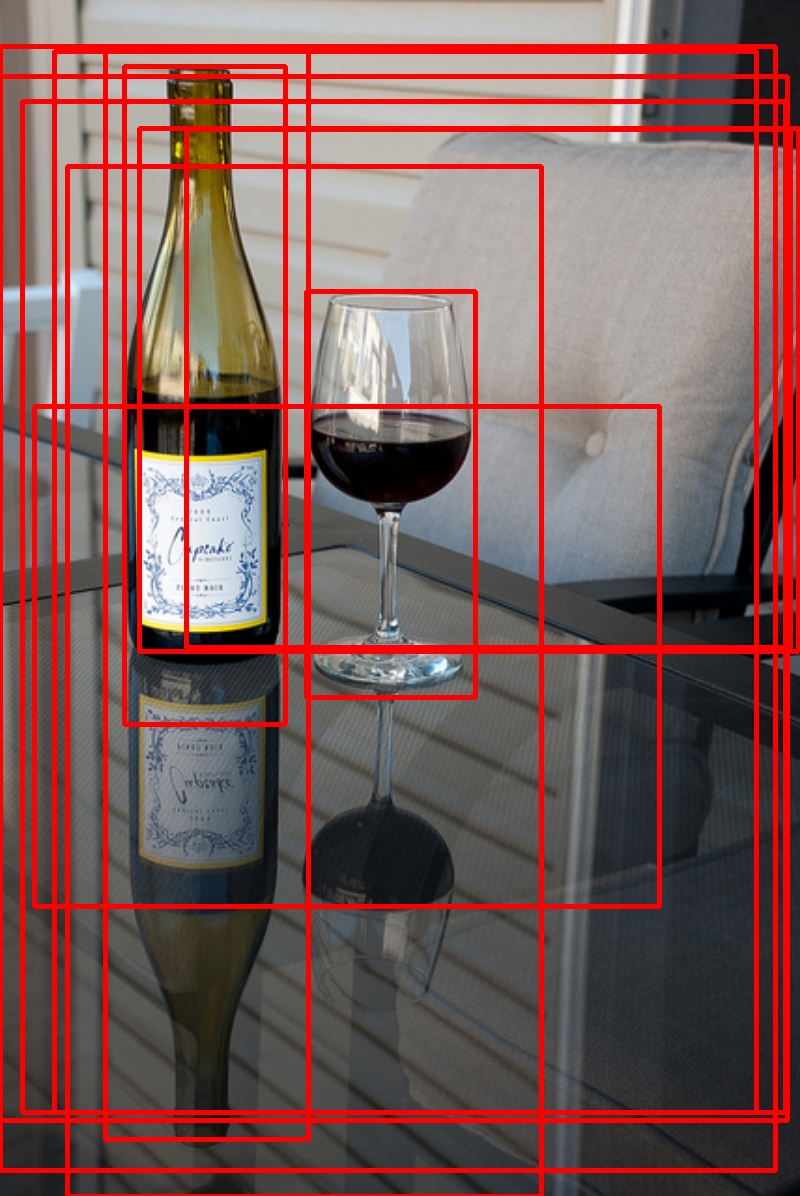}
        \caption{$0$}
        \label{fig:exp_ablation_threshold_0_0}
    \end{subfigure}\hspace{0.02in}
    \begin{subfigure}{.08\textwidth}
        \includegraphics[width=\textwidth]{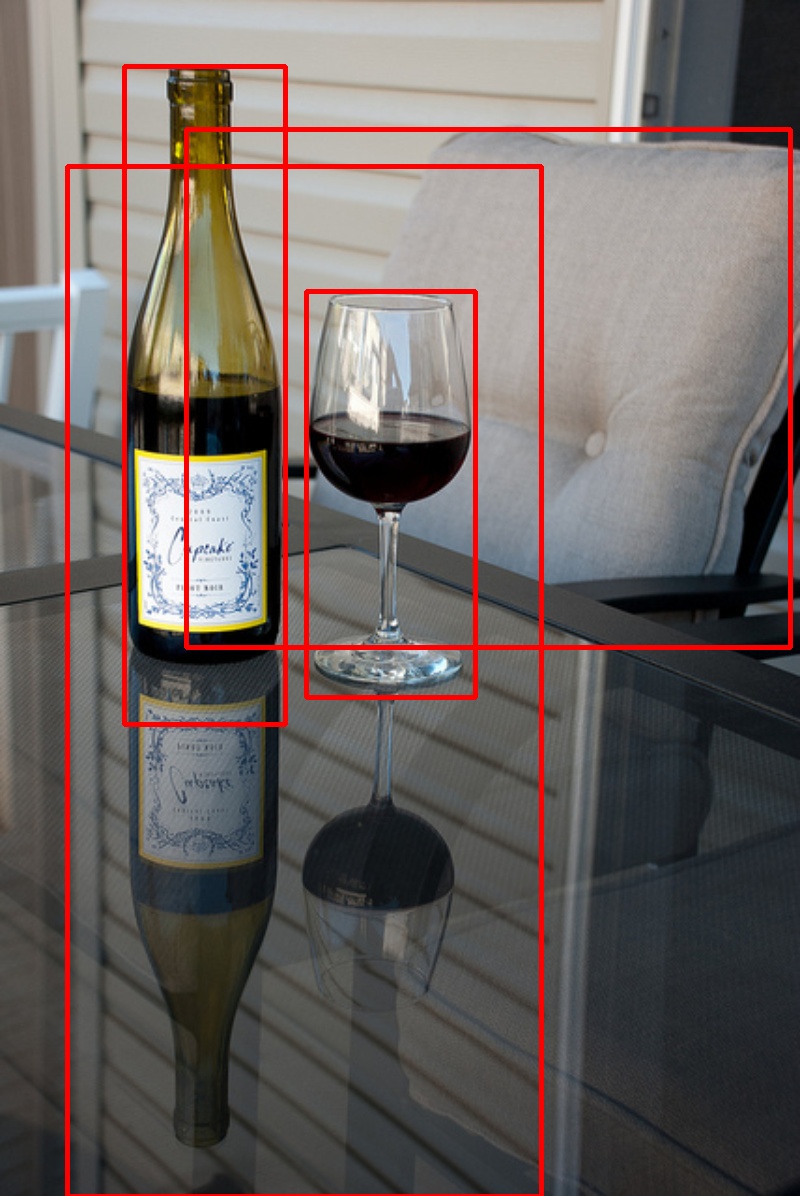}
        \caption{$0.3$}
        \label{fig:exp_ablation_threshold_0_3}
    \end{subfigure}\hspace{0.02in}
    \begin{subfigure}{.08\textwidth}
        \scalebox{-1}[1]{\includegraphics[width=\textwidth]{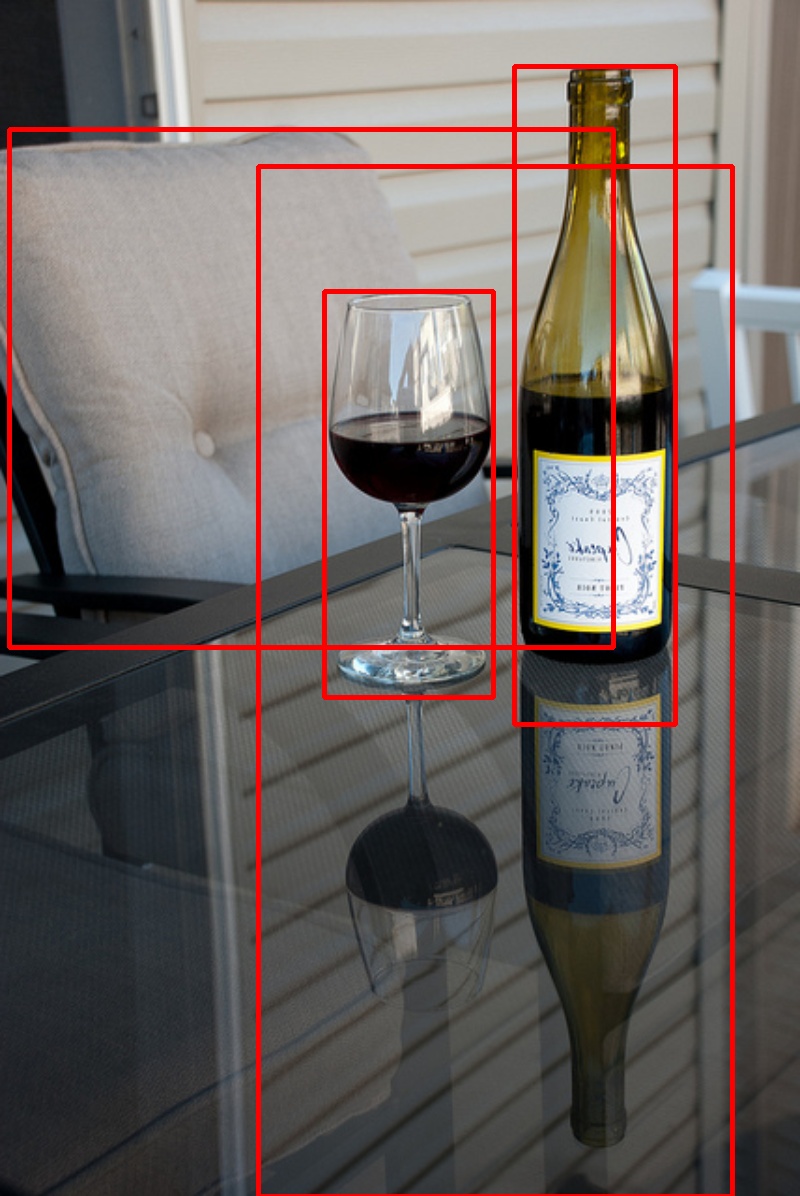}}
        \caption{$0.5$}
        \label{fig:exp_ablation_threshold_0_5}
    \end{subfigure}\hspace{0.02in}
    \begin{subfigure}{.08\textwidth}
        \scalebox{-1}[1]{\includegraphics[width=\textwidth]{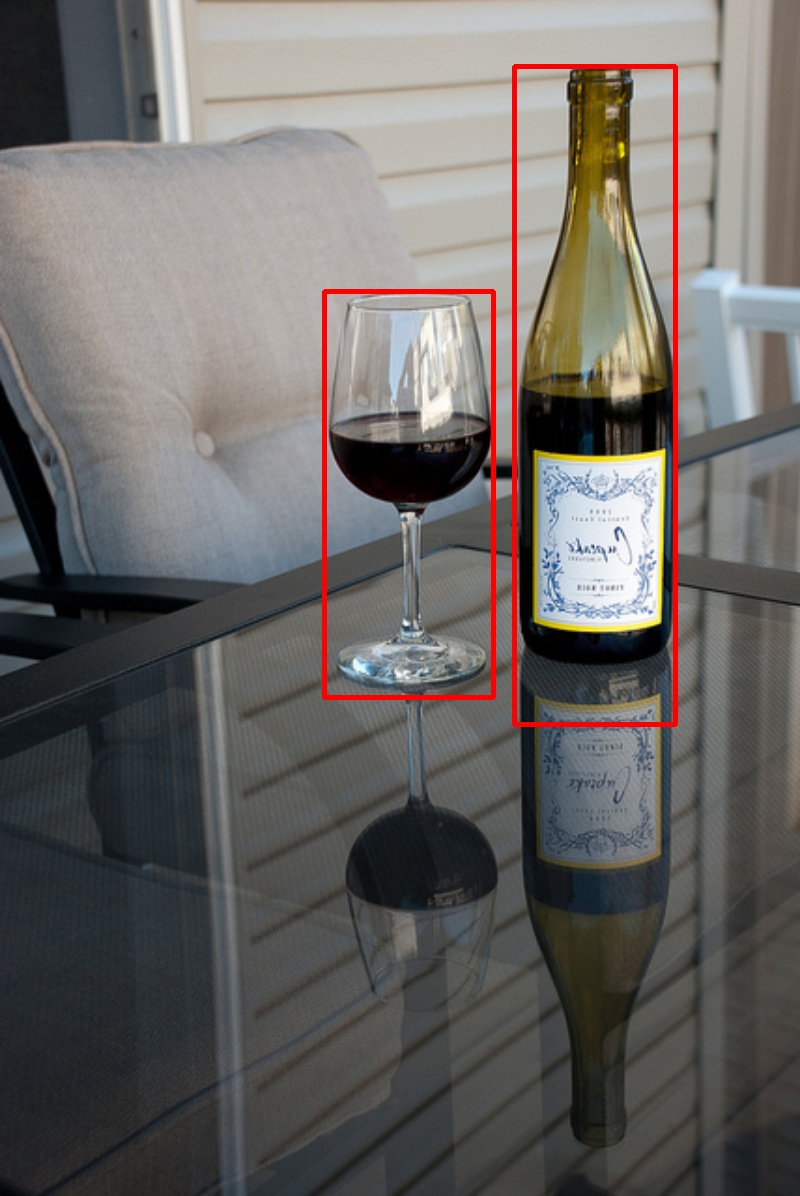}}
        \caption{$0.7$}
        \label{fig:exp_ablation_threshold_0_7}
    \end{subfigure}\hspace{0.02in}
    \begin{subfigure}{.08\textwidth}
        \includegraphics[width=\textwidth]{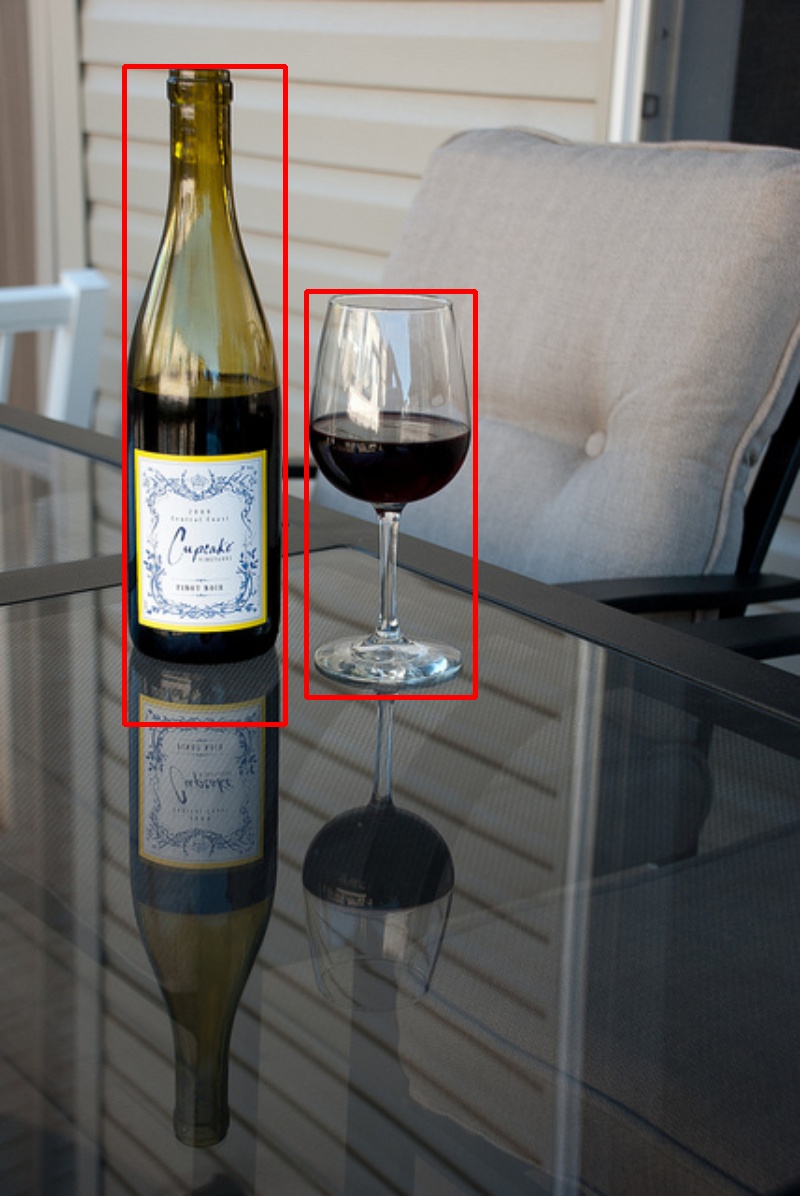}
        \caption{$0.9$}
        \label{fig:exp_ablation_threshold_0_9}
    \end{subfigure}
    \vspace{-0.1in}
    \caption{Visualization of predicted bounding boxes whose confidences are larger than $\tau$ for unlabeled data. Larger value of $\tau$ results in higher precision (e.g., remaining boxes after thresholding detect objects accurately), but lower recall (e.g., detected box at sofa is removed when $\tau\geq0.7$).}
    \label{fig:exp_ablation_threshold}
    \vspace{-0.1in}
\end{figure}

\begin{table*}[t]
    \centering
    \begin{tabular}{l|l|c|c|c|c}
        \toprule
        Protocol & { }Augmentation{ } & -- & \textsf{C} & { }\textsf{C+\{G,B\}}{ } & { }\textsf{C+\{G,B\}+Cutout}{} \\
        \midrule
        \multirow{2}{*}{5\% {\mscoco}{ }} & { }supervised{ } & { }18.67{ } & { }20.13{ } & { }20.78{ } & { }21.16{ } \\
        & { }{\ourmethod}{ } & { }24.49{ } & { }25.01{ } & { }24.70{ } & { }\textbf{25.12}{ } \\
        \midrule
        \multirow{2}{*}{10\% {\mscoco}{ }} & { }supervised{ } & { }24.05{ } & { }25.26{ } & { }25.92{ } & { }26.34{ } \\
        & { }{\ourmethod}{ } & { }\textbf{29.00}{ } & { }28.97{ } & { }28.41{ } & { }28.81{ } \\
        \bottomrule
    \end{tabular}
    \vspace{-0.05in}
    \caption{mAPs of supervised models and {\ourmethod} tested on a single fold of 5\% and 10\% protocols. We first train supervised models with different augmentation strategies (first row of each protocol), and pseudo labels generated form each supervised model are used to train {\ourmethod} models (second row of each protocol) accordingly.}
    \label{tab:ablation_pseudo_label}
    \vspace{-0.1in}
\end{table*}

\subsection{Size of Unlabeled Data}
\label{sec:exp_ablation_unlab}
\vspace{-0.03in}
While the importance of large-scale labeled data for supervised learning has been broadly studied and emphasized~\cite{deng2009imagenet,xiao2010sun,lin2014microsoft}, the importance on the scale of unlabeled data for semi-supervised learning has been often overlooked~\cite{oliver2018realistic}. 
In this study, we highlight the importance of large-scale unlabeled data in the context of semi-supervised object detector learning. We experiment with 5\% and 10\% labeled data of {\mscoco} while varying the amount of unlabeled data by 1, 2, 4, and 8 times more. 

The summary results are given in Table~\ref{tab:ablation_unlab}. While there still exists the improvement in mAPs when {\ourmethod} is trained with small amount of unlabeled data, the gain is less significant compared to that of supervised model with strong data augmentation. We observe clearly from Table~\ref{tab:ablation_unlab} that {\ourmethod} benefits from the larger amount of unlabeled training data. We make a similar observation from experiments on {\pascalvoc} in Table~\ref{tab:voc_result}, where the AP$^{0.5}$ of {\ourmethod} trained using trainval of {\vocb} as unlabeled data achieves 77.45, which is lower than that of supervised model with strong augmentations (78.21). On the other hand, {\ourmethod} trained with large amount of unlabeled data by combining {\vocb} and {\mscoco} achieves 79.08 AP$^{0.5}$.
This analysis may explain the slightly lower mAP of {\ourmethod} for 100\% protocol of {\mscoco} than that of the supervised model with strong data augmentation since the size of available unlabeled data is roughly the same as that of the labeled data.

\subsection{Hyperparameters $\lambda_u$ and $\tau$}
\label{sec:exp_ablation_hparams}
\vspace{-0.03in}
We study the impact of $\lambda_u$, a regularization coefficient for unsupervised loss, and $\tau$, the confidence threshold. Specifically, we test the {\ourmethod} with different values of $\lambda_u\,{\in}\,\{0.1,0.5,1,2,4\}$ and $\tau\,{\in}\,\{0,0.3,0.5,0.7,0.9\}$ on a single fold of 10\% protocol. The summary results are provided in Figure~\ref{fig:exp_ablation_hparams}. 
Firstly, the best performance of {\ourmethod} is obtained when $\lambda_u\,{=}\,2$ and $\tau\,{=}\,0.9$. We observe that the performance of {\ourmethod} deteriorates when $\lambda_u$ is too large (${>}\,2$) or too small (${<}\,0.5$), but it improves upon strong baseline consistently for $\lambda_u\,{\in}\,[1,2]$.
When there is no confidence-based box filtering, the gain of {\ourmethod}, if any, is marginal over the strong baseline. This is because lots of predicted boxes are indeed inaccurate, as shown in Figure~\ref{fig:exp_ablation_threshold_0_0}. Using larger value of $\tau$ allows to have pseudo box labels with higher precision (i.e., remaining boxes whose confidence is higher than $\tau$ are accurate), as in Figure~\ref{fig:exp_ablation_threshold_0_9}. However, if $\tau$ becomes too large, one would get a lower recall (e.g., bounding box at sofa in Figure~\ref{fig:exp_ablation_threshold_0_5} is filtered out in Figure~\ref{fig:exp_ablation_threshold_0_7}).
Figure~\ref{fig:exp_ablation_hparams} shows that the high precision (i.e., larger value of $\tau$) is preferred to high recall (i.e., smaller value of $\tau$) on 10\% protocol.

\subsection{Quality of Pseudo Labels}
\label{sec:exp_ablation_pseudo_label}
\vspace{-0.03in}
One intriguing question is whether the semi-supervised performance of the model improves with pseudo labels of higher quality. 
To validate the hypothesis, we train two additional {\ourmethod} models for 10\% protocol, where models are provided pseudo labels predicted by two different supervised models trained with 5\% and 100\% labeled data, whose mAPs are 18.67 and 37.63, respectively. Note that the {\ourmethod} on 10\% protocol achieves 29.00 mAP. {\ourmethod} trained with less accurate pseudo labels achieves only 24.25 mAP, while the one with more accurate pseudo labels achieves 30.30 mAP, confirming the importance of pseudo label quality. 

Inspired by this observation, we increase the augmentation strength to train the teacher model in order to get better pseudo labels, expecting a further improvement for {\ourmethod}. To this end, we train {\ourmethod} using different sets of pseudo labels that are provided by the supervised models trained with different data augmentation schemes. As in Table~\ref{tab:ablation_pseudo_label}, the performance of supervised models vary from mAP of 18.67 to 21.16 with 5\% labeled data and from 24.05 to 26.34 with 10\% labeled data. 
We observe an improvement in mAP by using more accurate pseudo labels on 5\% protocol, but the gain is not as substantial. We also do not observe a clear correlation between the accuracy of pseudo label and the performance of {\ourmethod} on 10\% protocol. While {\ourmethod} brings a significant gain in mAP using pseudo labels, our results suggest that the incremental improvement on the quality of pseudo labels may not bring in a significant extra benefit.

\section{Discussion and Conclusion}
\label{sec:conclusion}
While SSL for classification has made significant strides, label-efficient training for for tasks requiring high labeling cost is demanding.
We propose a simple (introducing only two hyperparameters that are easy to tune) and effective ($2{\times}$ label efficiency in low-label regime) SSL framework for object detection by leveraging lessons from SSL methods for classification. 
The simplicity of our method will provide a flexibility for further development towards solving SSL for object detection.

The proposed framework is amenable to many variations, including using soft labels for classification loss, other detector frameworks than Faster RCNN, and other data augmentation strategies. 
While {\ourmethod} demonstrates an impressive performance gain already without taking confirmation bias~\cite{zhang2016enhanced,arazo2019pseudo} issue into account, it could be problematic when using a detection framework with a stronger form of hard negative mining~\cite{shrivastava2016training,lin2017focal} because noisy pseudo labels can be overly-used. Further investigation in learning with noisy labels, confidence calibration, and uncertainty estimation in the context of object detection are few important topics to further enhance the performance of SSL object detection.

{\small
\bibliographystyle{ieee_fullname}
\bibliography{egbib}
}

\clearpage
\onecolumn
\appendix
\renewcommand{\thefigure}{A\arabic{figure}}
\setcounter{table}{0}
\renewcommand{\thetable}{A\arabic{table}}

\section{Learning Schedules}
\label{sec:setting}
In this section, we provide complete descriptions on different learning schedules used in our experiments. Note that the schedule \textsf{VOC} is only used for experiments related to {\pascalvoc}. Besides specified below, we adopt the learning settings as follows: \url{https://github.com/tensorpack/tensorpack/blob/master/examples/FasterRCNN/config.py}.

\subsection{\textsf{Quick}}
\begin{itemize}
    \item[$\bullet$] \textbf{LR Decay}: \\
    $\big[0.01 \,({\leq} 120k), 0.001 \,({\leq} 160k), 0.0001 \,({\leq} 180k)\big]$
    \item[$\bullet$] \textbf{Data processing}: Short edge size is sampled between 500 and 800 if the long edge is less than 1024 after resizing. 
    \item[$\bullet$] \textbf{Batch per image for training Faster RCNN head}: 64
\end{itemize}

\subsection{\textsf{Standard, $[n]\times$}}
\begin{itemize}
    \item[$\bullet$] \textbf{LR Decay}:\\
    $\big[0.01 \,({\leq} 120k), 0.001 \,({\leq} 160k), 0.0001 \,({\leq} 180k)\big]$
    \item[$\bullet$] \textbf{LR Decay ($2\times$)}: \\
    $\big[0.01 \,({\leq} 240k), 0.001 \,({\leq} 320k), 0.0001 \,({\leq} 360k)\big]$
    \item[$\bullet$] \textbf{LR Decay ($3\times$)}: \\
    $\big[0.01 \,({\leq} 420k), 0.001 \,({\leq} 500k), 0.0001 \,({\leq} 540k)\big]$
    \item[$\bullet$] \textbf{Data processing}: Short edge size is fixed to 800 if the long edge is less than 1333 after resizing. 
    \item[$\bullet$] \textbf{Batch per image for training Faster RCNN head}: 512
\end{itemize}

\subsection{\textsf{VOC}}
\begin{itemize}
    \item[$\bullet$] \textbf{LR Decay}:
     $\big[0.001 \,({\leq} 120k), 0.0005 \,({\leq} 160k)\big]$
    \item[$\bullet$] \textbf{Data processing}: Short edge size is fixed to 600 if the long edge is less than 1000 after resizing. Image is resized to have its longer edge to be 1000 if long edge is longer than 1000.
    \item[$\bullet$] \textbf{Batch per image for training Faster RCNN head}: 256
    \item[$\bullet$] \textbf{RPN Anchor Sizes}: $\big[8, 16, 32\big]$
\end{itemize}

\section{Data Augmentation in {\ourmethod}}
This section provides comprehensive results of Section~\ref{sec:exp_ablation_augmentation} to validate the importance of data augmentation in {\ourmethod}. In Table~\ref{tab:main_result_extended}, we provide two rows of results with {\ourmethod} (bottom) and the {\ourmethod} without strong data augmentation, i.e., ``Self-Training''. We observe significant gain in mAP on all cases, which validates the importance of the data augmentation in {\ourmethod}. 

\begin{table}[ht]
    \centering
    \begin{tabular}{l||c|c|c}
        \toprule
        Methods & { }5\% COCO{ } & { }10\% COCO{ } & { }100\%{ } \\
        \midrule
        {}Self-Training{ } & {21.80}$\pm$0.12 & {26.71}$\pm$0.27 & 37.57  \\
        {}{\ourmethod}{ } & \textbf{24.38}$\pm$0.12 & \textbf{28.64}$\pm$0.21 & \textbf{39.21} \\
        \bottomrule
    \end{tabular}
    \vspace{0.05in}
    \caption{Comparison in mAPs for different SSL methods on {\mscoco}. We report the mean and standard deviation over 5 data folds for 5\% and 10\% protocols. ``Self-Training'' refers to {\ourmethod} but without strong data augmentation on unlabeled data. We train {\ourmethod} with the strong augmentation for unlabeled data.}
    \label{tab:main_result_extended}
    \vspace{-0.1in}
\end{table}

\section{Extended Background: Unsupervised Loss in SSL}
\label{sec:ssl_loss}
In this section, we extend Section~\ref{sec:prelim_fixmatch} and provide unsupervised loss formulations for comprehensive list of SSL algorithms whose loss can be represented in Equation~\eqref{eq:unsup_loss_general}. 
For presentation clarity, let us reiterate definitions as follows:

\begin{equation}
    \ell_{u} = \sum_{x\in\mathcal{X}} w(x)\, \ell\left(q(x), p(x)\right)\label{eq:unsup_loss_general2}
\end{equation}
Here, we use $p(x)$ instead of $p(x;\theta)$ as in Equation~\eqref{eq:unsup_loss_general} for generality. Instead, let us denote $p(x; \theta)$ as a prediction of the model with parameters $\theta$ at training. 

Note that the unsupervised loss formulation of {\ourmethod} is following the form of Noisy Student (Section~\ref{sec:background_noisy_student}), which can be viewed as a combination of Self-Training (Section~\ref{sec:background_bootstrap}) and strong data augmentation. While we have shown such a simple formulation of {\ourmethod} brings in a significant performance gain at object detection, more complicated formulations (e.g., Mean Teacher (Section~\ref{sec:background_mean_teacher}) or MixMatch/ReMixMatch (Section~\ref{sec:background_mixmatch})) are amenable to be used in place of several design choices made for {\ourmethod}. Further investigation of {\ourmethod} variants is in the scope of the future work.

\subsection{Bootstrapping (a.k.a. Self-Training)~\cite{yarowsky1995unsupervised,mcclosky2006effective}}
\label{sec:background_bootstrap}
\begin{align}
\ell(q, p) &= \mathcal{H}(q,p) \\
w(x) &= 1 \mbox{ if }\max(p(x;\tilde{\theta}))\geq\tau \mbox{ else } 0\\
q(x) &= p(x; \tilde{\theta})\\
p(x) &= p(x; \theta)
\end{align}
where $\tilde{\theta}$ is the parameter of the existing model, which usually refers to a model trained on labeled data only until convergence.

\subsection{Entropy Minimization~\cite{grandvalet2005semi}}
\label{sec:background_entmin}
\begin{align}
\ell(q, p) &= \mathcal{H}(q,p) \\
w(x) &= 1\\
q(x) &= p(x; \theta)\\
p(x) &= p(x; \theta)
\end{align}
Note that gradient flows both to $q$ and $p$. To our best knowledge, Entropy Minimization is the only method that backpropagates the gradient through $q$.

\subsection{Pseudo Labeling~\cite{lee2013pseudo}}
\label{sec:background_pseudo_labeling}
\begin{align}
\ell(q, p) &= \mathcal{H}(q,p) \\
w(x) &= 1 \mbox{ if }\max(p(x;\theta))\geq\tau \mbox{ else } 0\\
q(x) &= \mbox{\textsc{one\_hot}}\left(\arg\max\left(p(x;\theta)\right)\right)\\
p(x) &= p(x; \theta)
\end{align}

\subsection{Temporal Ensembling~\cite{laine2016temporal}}
\label{sec:background_temporal_ensembling}
\begin{align}
\ell(q,p) &= \Vert q - p\Vert_2^2\\
w(x) &= 1\\
q^{(t)}(x) &= \alpha q^{(t-1)}(x) + (1-\alpha)p(x; \theta)\\
p(x) &= p(x; \theta)
\end{align}
We omit the ramp up and ramp down for $w(\cdot)$ in our formulation since it is dependent on the optimization framework. See \cite{laine2016temporal} for more details.

\subsection{Mean Teacher~\cite{tarvainen2017mean}}
\label{sec:background_mean_teacher}
\begin{align}
\ell(q,p) &= \Vert q - p\Vert_2^2\\
w(x) &= 1\\
q(x) &= p(x;\theta^{\text{EMA}}), \,\theta^{\text{EMA}}=\alpha\theta^{\text{EMA}}+(1-\alpha)\theta^{(t)}\\
p(x) &= p(x; \theta^{(t)})
\end{align}
We omit the ramp up and ramp down for $w(\cdot)$ in our formulation since it is dependent on the optimization framework. See \cite{tarvainen2017mean} for more details.

\subsection{Virtual Adversarial Training~\cite{miyato2018virtual}}
\begin{align}
\ell(q, p) &= \mathcal{H}(q,p) \\
w(x) &= 1\\
q(x) &= p(x;\theta)\\
p(x) &= p(\textsc{AP}(x); \theta),\,\mbox{\textsc{AP}($\cdot$): adversarial perturbation}
\end{align}

\subsection{Unsupervised Data Augmentation (UDA)~\cite{xie2019unsupervised}}
\label{sec:background_uda}
UDA uses a weak ($\alpha(\cdot)$), such as translation and horizontal flip, to generate a pseudo label, and strong augmentation ($\mathcal{A}(\cdot)$), such as RandAugment~\cite{cubuk2019randaugment} followed by Cutout~\cite{devries2017improved}, for model training.

\begin{align}
\ell(q, p) &= \mathcal{H}(q,p) \\
w(x) &= 1 \mbox{ if }\max(p(\alpha(x);\theta))\geq\tau \mbox{ else } 0\\
q(x) &\propto p(\alpha(x);\theta)^{\frac{1}{T}}\\
p(x) &= p(\mathcal{A}(x); \theta)
\end{align}

\subsection{FixMatch~\cite{sohn2020fixmatch}}
\label{sec:background_fixmatch}
FixMatch also uses a weak ($\alpha(\cdot)$), such as translation and horizontal flip, to generate a pseudo label, and strong augmentation ($\mathcal{A}(\cdot)$), such as RandAugment~\cite{cubuk2019randaugment} or CTAugment~\cite{berthelot2019remixmatch} followed by Cutout~\cite{devries2017improved}, for model training.

\begin{align}
\ell(q, p) &= \mathcal{H}(q,p) \\
w(x) &= 1 \mbox{ if }\max(p(\alpha(x);\theta))\geq\tau \mbox{ else } 0\\
q(x) &= \mbox{\textsc{one\_hot}}(\arg\max\left(p(\alpha(x);\theta\right)))\\
p(x) &= p(\mathcal{A}(x); \theta)
\end{align}

\subsection{Noisy Student~\cite{xie2019self}}
\label{sec:background_noisy_student}
\begin{align}
\ell(q, p) &= \mathcal{H}(q,p) \\
w(x) &= 1 \mbox{ if }\max(p(x;\tilde{\theta}))\geq\tau \mbox{ else } 0\\
q(x) &= \mbox{\textsc{one\_hot}}(\arg\max(p(x;\tilde{\theta}))))\\
p(x) &= p(\mathcal{A}(x); \theta)
\end{align}
where $\tilde{\theta}$ is the parameter of the model that is trained on labeled data only until convergence. In addition, Noisy Student perform data balancing across classes, which is not reflected in this formulation.

\subsection{MixMatch~\cite{berthelot2019mixmatch}}
\label{sec:background_mixmatch}
Note that MixMatch uses MixUp~\cite{zhang2017mixup} for unsupervised loss. It uses weak augmentation $\alpha(\cdot)$, such as translation and horizontal flip. 
\begin{align}
\ell(q, p) &= \mathcal{H}(q, p) \\
w(x) &= 1\\
\tilde{q}(x) &\propto \mathbb{E}_{\alpha}\big[p(\alpha(x);\theta)\big]^{\frac{1}{T}}\\
\tilde{q}(z) &\propto \mathbb{E}_{\alpha}\big[p(\alpha(z);\theta)\big]^{\frac{1}{T}}\\
q(x) &= \beta\tilde{q}(x) + (1-\beta)\tilde{q}(z)\\
p(x) &= p(\beta\alpha(x)+(1-\beta)\alpha(z); \theta)
\end{align}
where $x$ and $z$ are unlabeled data and $\beta$ is drawn from Beta distribution. While we present MixUp only between unlabeled data for presentation clarity, one may apply MixUp between labeled (with ground-truth label for $\tilde{q}$) and unlabeled data as well~\cite{berthelot2019mixmatch}.

\subsection{ReMixMatch~\cite{berthelot2019remixmatch}}
\label{sec:background_remixmatch}
Note that ReMixMatch uses MixUp~\cite{zhang2017mixup} for unsupervised loss. It also uses weak augmentation $\alpha(\cdot)$, such as translation and horizontal flip, and strong augmentation $\mathcal{A}(\cdot)$, such as CTAugment~\cite{berthelot2019remixmatch}.
\begin{align}
\ell(q, p) &= \mathbb{E}_{\mathcal{A}}\big[\mathcal{H}(q, p)\big] \\
w(x) &= 1\\
\tilde{q}(x) &\propto p(\alpha(x);\theta)^{\frac{1}{T}}\\
\tilde{q}(z) &\propto p(\alpha(z);\theta)^{\frac{1}{T}}\\
q(x) &= \beta\tilde{q}(x) + (1-\beta)\tilde{q}(z)\\
p(x) &= p(\beta\mathcal{A}(x)+(1-\beta)\mathcal{A}(z); \theta)
\end{align}
where $x$ and $z$ are unlabeled data and $\beta$ is drawn from Beta distribution. While we present MixUp only between unlabeled data for presentation clarity, one may apply MixUp between labeled (with ground-truth label for $\tilde{q}$) and unlabeled data as well~\cite{berthelot2019remixmatch}.

\end{document}